\documentclass[journal,onecolumn]{IEEEtran}

\usepackage{lineno,hyperref}
\modulolinenumbers[5]
\usepackage{cite}
\usepackage{amsmath,amssymb,amsfonts}
\usepackage{graphicx}
\usepackage{textcomp}
\usepackage{multirow}
\usepackage{caption}
\usepackage{blindtext}
\usepackage{xcolor}
\usepackage[utf8]{inputenc}
\usepackage[T1]{fontenc} 
\usepackage{mathtools}
\usepackage{tabularx}
\usepackage{subfigure}
\usepackage{dblfloatfix}
\usepackage[justification=centering]{caption}

\usepackage{multicol}
\usepackage{lipsum}
\usepackage{mwe}
\usepackage{longtable}
\usepackage[normalem]{ulem}
\usepackage{algorithm,algpseudocode}
\usepackage{varwidth}
\usepackage{amssymb}
\usepackage{lineno}
\usepackage{algorithm}
\usepackage{algorithmicx}
\usepackage{algpseudocode}
\usepackage{booktabs}
\usepackage{color}
\usepackage{xcolor}
\usepackage{tikz}
\usepackage{capt-of}
\usepackage{hyperref}
\usepackage{comment}
\usetikzlibrary{positioning, shapes.multipart}

\begin{document}

\title{A Blockchain Solution for Decentralized Machine Learning for IoT}
\author{Carlos Beis-Penedo, Francisco Troncoso-Pastoriza,  Rebeca P. D\'iaz-Redondo, Ana Fern\'andez-Vilas,  Manuel Fern\'andez-Veiga, Martín González-Soto
\thanks{Carlos Beis-Penedo (cbeis@det.uvigo.es), Rebeca P. D\'iaz-Redondo (rebeca@det.uvigo.es), Ana Fern\'andez-Vilas (avilas@det.uvigo.es) and Manuel Fern\'andez-Veiga (mveiga@det.uvigo.es) are with atlanTTic, ICLAB,  Escuela de Ingeniería de Telecomunicación. Campus universitario de Vigo, 36310, Spain.}
\thanks{Francisco Troncoso-Pastoriza (ftroncoso@cud.uvigo.es) is with Centro Universitario de la Defensa en la Escuela Naval Militar, Plaza de España, Marín, 36920, Spain.}
\thanks{Martín González Soto (mgsoto@gradiant.org) is with Centro Tecnolóxico de Telecomunicacións de Galicia (GRADIANT), Vigo, 36214, Spain.}
}

\maketitle

\abstract{The rapid growth of Internet of Things (IoT) devices and applications has led to an increased demand for advanced analytics and machine learning techniques capable of handling the challenges associated with data privacy, security, and scalability. Federated learning (FL) and blockchain technologies have emerged as promising approaches to address these challenges by enabling decentralized, secure, and privacy-preserving model training on distributed data sources. In this paper, we present a novel IoT solution that combines the incremental learning vector quantization algorithm (XuILVQ) with Ethereum blockchain technology to facilitate secure and efficient data sharing, model training, and prototype storage in a distributed environment. Our proposed architecture addresses the shortcomings of existing blockchain-based FL solutions by reducing computational and communication overheads while maintaining data privacy and security. We assess the performance of our system through a series of experiments, showing its potential to enhance the accuracy and efficiency of machine learning tasks in IoT settings.}


\section{Introduction}
\label{sec:intro}

The proliferation of Internet of Things (IoT) devices and applications has generated massive amounts of data that require advanced analytics and machine learning techniques for meaningful insights~\cite{HADDADPAJOUH2021100129}. IoT is impacting a wide range of industries, including smart cities, healthcare, industrial automation or transportation, by enabling real-time monitoring and control capabilities. However, traditional centralized machine learning models face challenges such as data privacy, security, and scalability. Federated learning (FL)~\cite{lim2020federated} is an emerging technique that addresses these challenges by enabling decentralized model training on distributed data sources while preserving data privacy and security. Despite its promise, FL still faces several technical challenges such as non-iid data distribution, communication overhead, and straggler nodes~\cite{LiTian2020}.

In the traditional FL approach, multiple devices work together to train a machine learning model. Each node keeps its own data set that is not shared, thus, data resides on trusted nodes. This scenario is particularly convenient for IoT applications, where devices often generate sensitive data that must be protected from unauthorized access. Each node locally trains its own model, which is later shared with the other nodes within the FL setting. These local models are aggregated to build up a global model without exposing the local data. However, this exchange of model updates introduces new security and privacy concerns~\cite{lyu2022privacy}. Some of the most widely known security challenges are related to protect the FL setting against the following attacks: (i) data poisoning attacks~\cite{tolpegin2020data}, where malicious nodes inject corrupted or misleading data into the training process, compromising the accuracy of the global model; (ii) model inversion attacks~\cite{wang2022protect}, where adversaries aim to reconstruct individual data samples from aggregated model updates, potentially revealing sensitive information; (iii) sibyl attacks~\cite{fung2018mitigating,AdversarialLens}, where malicious entities create multiple fake nodes to disproportionately influence the federated learning process; and (iv) collusion attacks~\cite{lyu2023poisoning}, which involve a group of malicious nodes conspiring to manipulate the global model.


To address these challenges, recent research has proposed FL solutions that implement blockchain technology for secure and efficient data sharing, model training, and prototype storage in a distributed environment. Blockchain technology~\cite{Christidis16}, by providing a tamper-proof distributed ledger for storing and sharing data, models, and training results, enables collaboration among multiple parties without the need for a central authority, thereby significantly enhancing data privacy and security in the process.

Consequently, Blockchain technology can be applied to address these challenges by providing a secure and transparent platform that stores and manages the global model and model updates from each local node. Because of its inherent cryptographic security features and consensus mechanisms, it is possible to ensure the integrity and confidentiality of the models during the exchange process. This approach effectively mitigates the risks associated with any potential threats to the privacy and security of federated learning models~\cite{Rahman20}. 

In light of these ongoing challenges, we introduce a novel IoT solution that combines blockchain technology and federated learning for decentralized and collaborative machine learning. The proposed system consists of two primary elements that collaborate to create a secure, efficient, and cooperative machine learning environment: first, the network of sensors, which is accountable for gathering data from a variety of sources within the IoT ecosystem; and second, the nodes that process the data, which is captured and transmitted from the sensor network.

Since our approach is intended to be used in IoT scenarios, local nodes work using an incremental algorithm optimized to work in resource-constrained environments (memory and computation). Thus, we  employ the XuILVQ federated learning algorithm~\cite{Martin}, where each local node trains a local model using its own data and shares only the model (prototype) updates with the other ones, enhancing data privacy while allowing the global model to learn from all nodes. To improve the system's security and privacy, we incorporate secure data aggregation for data transmission between sensors and nodes, and Ethereum blockchain as a secure and transparent platform for storing and managing the global model and model updates from each node. We have performed several experiments in order to assess the suitability (accuracy, security and efficiency tests) for IoT scenarios using resource-constrained devices.

Therefore, we propose a solution that combines XuILVQ federated algorithm with the Ethereum blockchain technology to assure secure and efficient data sharing, model training and prototype storage (intermediate models) in a distributed FL setting. An important aspect of our proposal is that this solution reduces computational and communication overheads, with the consequent energy savings, while prioritizing data privacy and security.



The document is structured as follows. Section~\ref{ref:relatedWork} provides an overview of previous research in the field of Federated Learning (FL) and Blockchain (BC) technology applied to Internet of Things (IoT) applications. The section categorizes and reviews various research works, highlighting their contributions to the field and the challenges they address. Section~\ref{sec:background} provides an introduction to the key concepts and technologies used in the paper, including FL and Distributed Ledger Technology (DLT) such as blockchain. The section explains how these technologies work and how they can be applied to address the challenges of collaborative machine learning in IoT environments. Section~\ref{sec:system} presents the proposed architecture, detailing its components and their interactions. Section~\ref{sec:results} includes the results of various tests, analyzing the performance of the system. Finally, Section~\ref{sec:con} draws conclusions based on the findings, highlighting the implications of this research for the field of collaborative machine learning in IoT and suggesting potential directions for future work.

\section{Related Work}
\label{ref:relatedWork}
In recent years, the integration of Federated Learning  (FL) and Blockchain (BC) technologies has gained significant attention, particularly in enhancing security and efficiency in Internet of Things (IoT) environments~\cite{GILL2022100514} . As Federated Learning aims to train a high-quality centralized model without exposing raw data across numerous clients \cite{konečný2017federated}, it faces inherent challenges such as unreliable network situations and intermittent client availability, which can complicate the training process and threaten data privacy and model integrity \cite{eswaFL, AdversarialLens}.

Blockchain technology offers a promising solution to these challenges by decentralizing the training process, thereby enhancing the security and robustness of Federated Learning systems. This approach mitigates risks such as data poisoning and model inversion attacks, leveraging blockchain's immutable and transparent nature to secure FL processes \cite{Kalapaaking20231703, Li20231286}.

Significant advancements have also been made in applying Blockchain to manage and secure communications and operations in IoT, with notable studies exploring new blockchain architectures and integrating trusted execution environments to improve security. However, the integration of these technologies in real-world applications remains complex~\cite{AHMAD2021100365}, with unresolved issues like non-IID data distribution, communication overhead, and node reliability.

In addressing these challenges, recent literature has proposed several innovative approaches. For instance, optimizing data flow to reduce latency in communications \cite{LiTian2020}, employing smart contracts to automate and secure interactions within IoT networks \cite{Christidis16}, or developing lightweight blockchain frameworks that are feasible for IoT devices, enhancing real-time data processing capabilities \cite{Rahman20}.

Further research has explored the efficiencies of various blockchain and federated learning configurations in IoT scenarios, assessing their applicability in environments requiring real-time processing \cite{Chen2023, 9448383}. Novel solutions like committee consensus models have been introduced to improve consensus efficiency and reduce communication overhead, which are critical in maintaining the real-time processing capabilities of IoT systems \cite{CommitteeConsensus}.

Comprehensive reviews have juxtaposed different approaches within the blockchain and FL domains, providing valuable insights into their application across industrial and consumer IoT devices \cite{9766205, 9892039}. Additionally, studies focusing on security enhancements in FL and BC remark the need to develop robust defenses against emerging security threats in these technologies \cite{securingFLreview}.

Building on this foundation, our research proposes integrating an incremental learning algorithm optimized for data privacy within a blockchain framework. This combination aims to leverage the strengths of both technologies, ensuring secure, efficient, and real-time data processing in IoT environments \cite{Martin}. This integrated approach is expected to address the operational challenges and enhance the applicability of FL and BC in diverse real-world scenarios.
\section{Background}
\label{sec:background}

This section provides an overview of the four most relevant technologies that are used in our proposal to establish the baseline for the main contributions of our work, which are presented later.
In Section \ref{sec:background-fl} we describe the main characteristics of Federated Learning, while in Section \ref{sec:background-il} we detail the strategy behind the incremental learning algorithms. Then, in Section \ref{sec:background-ledger} the bases of Distributed Ledger Technology (DLT), such as blockchain, are summarized and, finally, Section \ref{sec:background-aggregation} includes some relevant aspects of the secure aggregation of data.

\subsection{Federated Learning}
\label{sec:background-fl}

Federated Learning (FL) is a machine learning strategy that enables multiple devices to collaboratively train a model without sharing their raw data~\cite{rieke2020future, sattler2020robust}. In this paradigm, devices maintain their data locally, compute model updates, and periodically share these updates (i) with other devices in the trusted network or (ii) with a central coordinating server or aggregator, which is in charge of combining the local models to obtain a global one that is later sent to all the nodes in the FL scheme. This process finishes when the global model converges. Thus, instead of having a centralized approach, where the data must be gathered in a central computation node, the decentralized philosophy of FL intends to preserve privacy and security, as the raw data never leaves the local devices.

The use of a a central coordinating server or aggregator offers some advantages. First, it simplifies the process of model training and update aggregation, as all updates are gathered and combined by a central entity. This makes the model training process more manageable and efficient, especially in large-scale federations involving many clients. Second, the central server is also a trusted entity for model validation and consistency checks, ensuring the quality and integrity of the global model. Finally, this central element can also facilitate communication and synchronization among clients, enhancing the overall efficiency of the federated learning process.


Federated Learning can be categorized into horizontal FL or vertical FL according to the data distribution and model training methodologies~\cite{yang2019federated}. In the former, all participating entities (or nodes) have the same feature space but different samples, which is broadly used in IoT settings~\cite{zhang2023federated}. The latter is used when different entities have different feature spaces from the same set of elements, and it is typically applied in the e-health field~\cite{liu2024vertical}. Additionally, FL settings can be classified into cross-silo federations~\cite{huang2022cross}, which involves organizations or entities with large computational resources, or cross-devices federations~\cite{karimireddy2021breaking}, which involves a large number of devices with smaller computational resources.




FL is particularly beneficial in sectors such as healthcare, telecommunications, and transportation where data privacy is key and collaboration for shared goals is required~\cite{li2020review}. In healthcare, for instance, FL allows multiple hospitals to collaboratively train models for disease prediction or treatment optimization, without sharing sensitive patient data. This preserves patient privacy while enabling the development of robust and effective models using diverse data sources. Similarly, in telecommunications and transportation, FL can enable collaborative model training across different network devices or vehicles for tasks like network optimization or autonomous driving, without compromising user or vehicle data privacy.

Although FL keeps data at each computation node, there would be 
privacy leaks because of the exchange of the AI model information. Therefore, different techniques have arisen to try to assure privacy in FL settings. Differential privacy (DP)~\cite{DifferentialPrivacy:CD06} adds noise to the data or model parameters during the training stage; Homomorphic encryption (EC)~\cite{HomomorphicEncryption:CCD17} performs computations on the encrypted data domain and Secure Multi-Party Computation (SMPC)~\cite{SecureMPC:EKR18} allows a group of nodes to collaborative compute a function, while keeping the data secret. 

Besides, when the FL setting has a central aggregator, there is also a new risk or weakness: it is a potential single point of failure. Therefore, the combination of centralized coordination and decentralized data storage and computation entails a challenge: ensuring the robustness and resilience of the whole FL system. For this to be possible, it is essential to design mechanisms that can effectively mitigate this risk, balance efficiency and privacy, and enhance the overall security of the system.


Consequently, we explore how the advantages of FL can be further enhanced by integrating a distributed ledger technology, which adds a layer of security and trust to the system. This integration would make FL systems more robust and trustworthy, opening up new possibilities for secure, privacy-preserving collaborative machine learning in various applications.

\subsection{Incremental Learning Algorithms}
\label{sec:background-il}

Incremental learning is a strategy in machine learning where the model has the ability to learn and improve over time with the arrival of new data. Unlike the traditional batch learning methods, where the model is trained at once, incremental learning sequentially trains the model on data instances. This approach is particularly useful in scenarios where data arrives in streams, the entire dataset is too large to fit into memory or the computation devices have severe resource restrictions.

The underlying philosophy is that the learning process is never considered to be finished. Thus, new data means new knowledge to be acquired in the model in an iterative learning process, which assures the model would adapt to new patterns and changes in the data distribution over time. This is also especially convenient in IoT settings, where devices usually do not have high memory and computational resources. 


One particular algorithm that utilizes this strategy is Incremental Learning Vector Quantization (XuILVQ)~\cite{Martin}, which we use in our proposed architecture. XuILVQ is an enhanced version of the Learning Vector Quantization (LVQ)~\cite{LVQ} algorithm, which is a prototype-based supervised classification algorithm. LVQ is known for its simplicity and effectiveness in handling high-dimensional data.

In the context of LVQ, a prototype is a representative example of a class. During the training process, these prototypes are iteratively adjusted to be representative of all the identified classes. XuILVQ improves LVQ by incorporating a nearest neighbour rule in the adjustment process, which results in a more accurate and efficient prototype set.

XuILVQ is suitable for our IoT FL setting where each node receives new data in sequence and, consequently, it is iteratively processed. The model (set of prototypes) is progressively updated and shared with the collaborative network~\cite{gonzalez2024decentralized}, ensuring and adaptive and efficient learning process. 

We propose to add a Distributed Ledger Technology (DLT) layer over these collaborative FL scheme, which would increase security and transparency to the prototype sharing process.


\subsection{Distributed Ledger Technology}
\label{sec:background-ledger}


Distributed Ledger Technology (DLT), such as blockchain, is a decentralized database system that allows multiple parties to share, verify, and store data securely without a central authority~\cite{siano2019survey}. The data is stored in a chain of blocks, each containing a set of transactions or records. Every new block added to the chain is cryptographically linked to the previous block, ensuring the immutability and tamper-proof nature of the stored data. DLT is based on consensus mechanisms, such as Proof of Work or Proof of Stake, that validate transactions and add new blocks to the chain. 

The decentralized nature of DLT is suitable for any application where multiple parties need to maintain consistent records without relying on a central authority. This feature can enhance its resilience against single points of failure and promote a more equitable distribution of control. Once data is written to a blockchain, it cannot be retroactively altered, which is essential in contexts where traceability and auditability of data are paramount. Thus, these transparent and verifiable transactions reduce concerns about data misuse, which helps in settings where trust and data integrity are required.

Consequently, DLT technologies are used in different application domains, such as in the medical field~\cite{kuo2017blockchain}, where they help ensuring confidentiality on transactions of medical records (keeping and sharing records among different healthcare organizations, for instance). The public sector is another field where DLT is key to assure transparency~\cite{olnes2017blockchain} or in peer-to-peer energy trading in microgrids~\cite{zia2020microgrid}, where blockchain can secure the transactions and promotes an efficient energy management.


\subsection{Secure Data Aggregation}
\label{sec:background-aggregation}

Ensuring secure data aggregation is key in distributed systems, especially when working with sensitive data. Thus, maintain the confidentiality and accuracy of the aggregated data is the first goal to achieve, even under potential adversarial conditions. The SwiftAgg+ protocol~\cite{jahaninezhad2022swiftagg} arises as an innovative solution for this issue.

SwiftAgg+ works as follows. First, each computation element (a sensor which is gathering data, for instance) segments the data using the 
Shamir's Secret Sharing scheme~\cite{shamir1979share}. Since this scheme distributes a secret among multiple elements without revealing it to any individual participant, SwiftAgg+ ensures no single node in the network has access to the complete data originating from a sensor, reinforcing the security and privacy of the data. After that, and once the data is shared, each node applies a Maximum Distance Separable (MDS) approach ~\cite{kasami1984probability} to generate redundant data shares. This enables the recovery of the original dataset even if some data are lost or compromised.




\section{Proposed Architecture}
\label{sec:system}

\begin{figure}[h]
  \centering
  \includegraphics[width=0.46\textwidth]{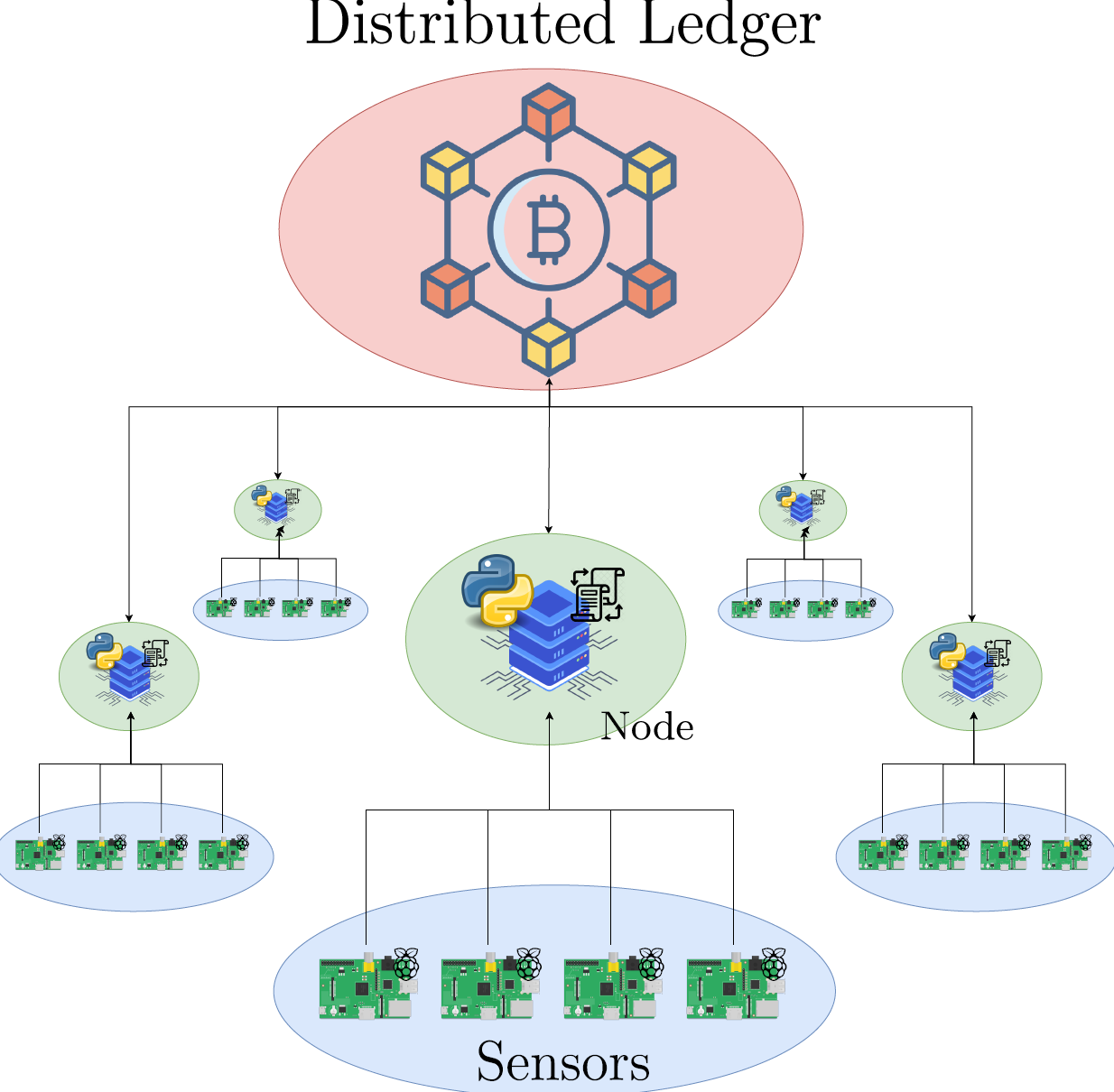} 
  \caption{Network showcase of the system.}
  \label{fig:network}
\end{figure}

\begin{figure*}[t]
  \centering
  \includegraphics[width=\textwidth]{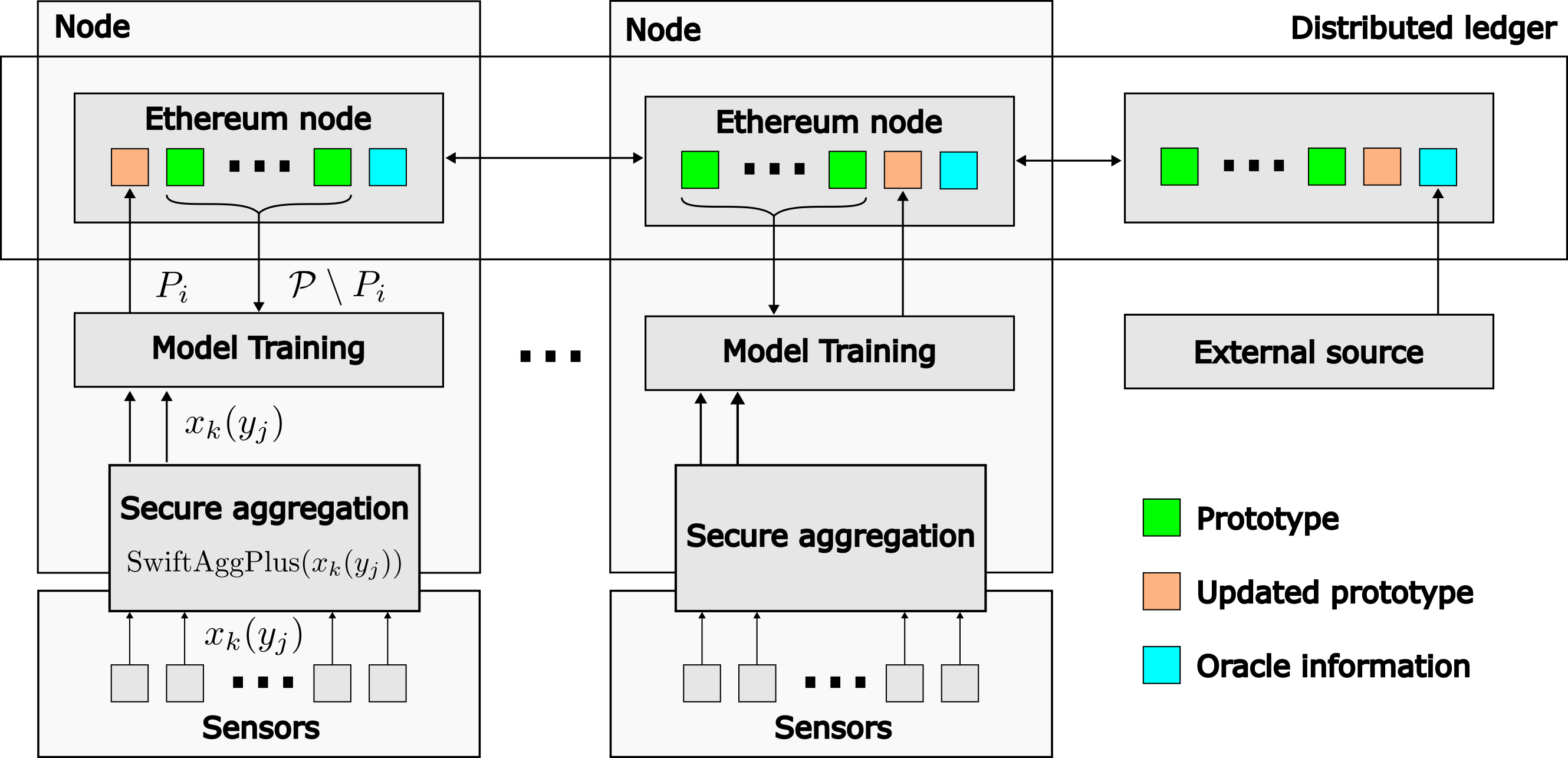} 
  \caption{Logical model of the system.}
  \label{fig:Logical Model}
\end{figure*}

The proposed architecture, illustrated in Figure \ref{fig:network}, is designed as a three-tier hierarchical structure, strategically crafted to effectively address the challenges of collaborative machine learning in IoT settings. The lowest tier of the architecture is composed of sensors, which are in charge of gathering data and, after that, sharing it with its corresponding computing node. The middle layer is precisely composed of a set of computing nodes, in charge of processing the received information. Finally, the distributed ledger is on the top of the architecture, being its highest tier.

Sensors are in charge of gathering data in batches of samples. Each batch is later shared with their corresponding node using the 
SwiftAgg+ protocol to ensure secure aggregation. Sensors are denoted as $x_k$, for $k = 1, \dots, s$, and each batch is denoted as $y_j$. Thus, $x_k(y_j)$ denotes the data gathered by node $x_k$, as depicted in Figure~\ref{fig:Logical Model}.

Consequently, each node is in charge of using the gathered data from its sensors to train its local model using the incremental learning vector quantization algorithm (XuILVQ), as detailed in Section \ref{sec:background-il}. Since XuILVQ is a a prototype-based algorithm, the local model is already defined by the set of prototypes, which are denoted as $P_i$ in Figure~\ref{fig:Logical Model}.

Finally, the distributed ledger uses the Ethereum technology, as detailed in Section \ref{sec:background-ledger}, to have a distributed database built with the information kept by each node. Thus, each node, as detailed in Figure~\ref{fig:Logical Model}, has already three tasks or components: the SwiftAgg+ technology for secure aggregation, the XuILVQ algorithm for decentralized learning and an Ethereum part for ensure the immutable, transparent, and secure record of information.




\subsection{Algorithmic approach}
\label{sec:algorithmic_approach}

After the overview of the proposed architecture, in this section we detail the two main algorithms used for this decentralized learning process: the first one focuses on the training tasks carried out locally by each node (Algorithm~\ref{alg:nodeFlow}) and the second focuses on the tasks carried out by each node to collaboratively train the global model (Algorithm~\ref{alg:trainModel}). Table~\ref{tab:notation1} summarizes the notation used in both algorithms, where each entry details a parameter or variable, its mathematical notation, and a brief description of its role in the learning procedure.



The local training performed by each node is detailed in Algorithm~\ref{alg:nodeFlow}. First, each node gathers the data provided by its set of sensors ($\mathcal{S_n}$) creating a data set $\mathcal{A}$ using the $SwiftAggPlus()$ function to assure a secure aggregation, which is combined into the set $\mathcal{D}$. Each node uses the \_seeAll() function (Algorithm~\ref{alg:seeAll}) to fetch the set of all prototypes $ P_{-n} $ from the other nodes in the network, excluding its own prototypes $ P_n $. After that, the node trains the model with  $\mathcal{D} $ and the prototypes from other nodes $ P_{-n} $. This process yields an updated prototype set $ P_n' $ and an updated local model $ M_n' $. 

Algorithm~\ref{alg:trainModel} gives a more detailed perspective of the local training model $M$ given a pair of data sample and label $(x,y)$, respectively. If the number of elements in the training buffer $\mathcal{B}$ is less than or equal to $2$, the pair is directly added to $\mathcal{B}$. This procedure ensures that the model is initially composed of a diverse set of data samples. Otherwise, the algorithm identifies the two nearest prototypes $s_1$ and $s_2$, along with their distances $d_{s_1}$ and $d_{s_2}$ from $x$, using the function $f\_nearest$. Then, it updates these prototypes based on $x$ and $y$. The decision to add a new prototype to the set $\mathcal{B}$ depends on whether the new data sample brings new information that is not properly represented by the existing prototypes. Specifically, if $y$ does not belong to the class of $B$, or if the distances $d_{s_1}$ or $d_{s_2}$ are greater than their respective thresholds $thr_{s_1}$ and $thr_{s_2}$, the pair $(x, y)$ is added to $B$. In all other scenarios, the prototype set $\mathcal{B}$ and prototype $s_1$ are updated based on $x$, $y$, and a parameter $m_{s_1}$ using the function update. This complex decision-making process ensures that the prototypes in $\mathcal{B}$ are always representative of the data samples and are continually refined as new data arrives.

Therefore, at each training iteration, the system may ingest data from two sources: (i) one drawn directly from the sensors, and (ii) another obtained from the exchange of prototypes among the nodes.

Finally, if appropriate, the new prototype set $P_n'$ is updated to the blockchain. This decision is key in the collaborative and decentralized learning process and it is based on a statistical test: if the difference between the average of the updated prototypes $ \bar{P_n'} $ and the average of the current prototypes $ \bar{P_n} $ exceeds a certain threshold, in our case, a 5\% performance interval, then $ P_n' $ is added to the blockchain via the \_addPrototype() function. Therefore, prototypes are evaluated based on their performance metrics over a reduced validation test set, and only those that achieve metrics over 5\% of the peak performance are considered appropriate to be shared and enrich the global learning model, that is, to be uploaded to the distributed ledger. This ensures only substantial updates are recorded on the blockchain, saving storage and computational resources.

\begin{table}[H]
\centering
\caption{Notations used in Algorithm~\ref{alg:nodeFlow} and Algorithm~\ref{alg:trainModel}}
\label{tab:notation1}
\begin{tabular}{|l|p{4.6cm}|}
\hline
\textbf{Algorithm~\ref{alg:nodeFlow}} & \textbf{Description} \\ \hline
$M$ & Training Model \\ \hline
$M'$ & Updated training model after processing \\ \hline
$\mathcal{P}$ & Set of prototypes \\ \hline
$\mathcal{N}$ & Set of nodes in the network \\ \hline
$\mathcal{S}_{n}$ & Set of sensors attached to node $n$ \\ \hline
$\mathcal{A}_s$ & Set of aggregated data obtained from the input samples for a given sensor $s$ \\ \hline
$\mathcal{D}$ & Combined dataset from all sensors attached to a node \\ \hline
$x$ & Individual data sample collected from sensors \\ \hline
$y$ & Label or class of the data sample \\ \hline
$P_{-n}$ & Set of prototypes from other nodes, excluding the current node \\ \hline
$\mathrm{SwiftAggPlus}(x)$ & Function to apply the SwiftAgg+ secure aggregation on sample $x$ \\ \hline
\end{tabular}
\vspace{0.5cm}

\begin{tabular}{|l|p{3.8cm}|}
\hline
\textbf{Algorithm~\ref{alg:trainModel}} & \textbf{Description} \\ \hline
$\mathcal{B}$ & Buffer holding data samples during the training process \\ \hline
$f\_nearest(B, x, n)$ & Function to find the $n$ nearest prototypes to $x$ in buffer $B$ \\ \hline
$s_1, s_2$ & Nearest prototypes to the data sample $x$ \\ \hline
$d_{s_1}, d_{s_2}$ & Distances from sample $x$ to prototypes $s_1$ and $s_2$, respectively \\ \hline
$thr_{s_1}, thr_{s_2}$ & Threshold distances for prototypes $s_1$ and $s_2$, used for prototype updates \\ \hline
$m_{s_1}$ & Parameter to update the prototype $s_1$ based on the new data \\ \hline
update($B$, $s_1$, $x$, $y$, $m_{s_1}$) & Function to update the prototype set $B$ using $x$, $y$, and $m_{s_1}$ \\ \hline
\end{tabular}
\end{table}

  \begin{algorithm}[t]
    \caption{Data Flow in Nodes}
    \label{alg:nodeFlow}
    \begin{algorithmic}[1]
      \State Initialize $M$ on each node.
      \For{node $n \in [N]$}
          \State Initialize empty set $\mathcal{A}$
          \For {each sensor {$s \in S_n$}}
            \State Aggregate new data samples $x$ using $\mathrm{SwiftAggPlus}(x)$
            \State Collect aggregated data in $\mathcal{A}_s$
          \EndFor
          \State Combine data from all sensors into set $\mathcal{D}$
          \State Fetch prototypes from other nodes as $P_{-n}$
          \State Update model and prototypes with train\_model(M,[$\mathcal{D}$ and $P_{-n}$])
          \If{Trust-Based selection passed on new model}
              \State Update blockchain with new prototypes
          \EndIf
      \EndFor
      \State \textbf{return} $M'$
    \end{algorithmic}
  \end{algorithm}

\begin{algorithm}[t]
\caption{train\_model($M$, [$x$, $y$])}
\label{alg:trainModel}
\begin{algorithmic}[1]
\If{less than 3 data samples in buffer $\mathcal{B}$}
    \State Add data sample $(x, y)$ to $\mathcal{B}$
\Else
    \State Identify nearest prototypes $s_1$ and $s_2$ and their distances $d_{s_1}$, $d_{s_2}$ using $f\_nearest(\mathcal{B}, x, 2)$
    \If{$y$ is not represented in $\mathcal{B}$ or if $d_{s_1} > thr_{s_1}$ or $d_{s_2} > thr_{s_2}$}
        \State Add $(x, y)$ to $\mathcal{B}$
    \Else
        \State Update prototype $s_1$ using update($\mathcal{B}$, $s_1$, $x$, $y$, $m_{s_1}$)
    \EndIf
\EndIf
\State \Return updated model $M$
\end{algorithmic}
\end{algorithm}


In summary, the proposed architecture integrates (i) XuILVQ for local model training, (ii) secure aggregation for exchange of data between sensors and nodes, and (iii) a distributed ledger for prototype sharing among nodes. This combination of techniques and technologies enables a robust, efficient, and secure framework for decentralized machine learning in IoT environments.




\subsection{Enhancing Blockchain Solutions with Oracle-Integrated Smart Contracts}
\label{sec:oracle}

The blockchain solution that we include in our proposal was built on the Geth platform~\cite{geth} and it incorporates two essential smart contracts, the PrototypeBuffer and the DataOracle, both developed in Solidity\footnote{https://soliditylang.org/}. These smart contracts manage prototypes and their associated external data, thereby enhancing the functionality and utility of our blockchain solution.

At this point, it is critical to clarify why we include an oracle \cite{caldarelli2020understanding} (see Figure \ref{fig:network}) and why we deem it essential for the whole solution. The idea of allowing an external input via an oracle comes from a practical consideration of real-world data integration into blockchain-based systems. We envision situations where a company might want to participate in the project but is unable or unwilling to grant direct access to their sensor data network. An oracle provides a feasible solution in this context, serving as a secure, single point of external data access. By fortifying this oracle, we can safeguard any data incoming from external sources, thereby enhancing the overall data security of our system.

The PrototypeBuffer smart contract operates as a management system for prototypes and their respective data updates facilitated through the oracle. Employing the OpenZeppelin library, it implements the ownership pattern (Ownable). This smart contract empowers prototype owners to control and update their prototype information, view information about all prototypes and their associated oracle data, and supports the addition of new owners and modification of the oracle address.

As depicted in Figure~\ref{fig:uml}, the PrototypeBuffer and DataOracle smart contracts encompass a suite of state variables, mappings, events, functions, and modifiers serving distinct roles in the proposed architecture. The PrototypeBuffer contract maintains information about the owners, their prototypes, and the oracle data, while the DataOracle contract shoulders the responsibility of providing and updating external data.

The PrototypeBuffer contract operates as a management system for prototypes, allowing owners to manage and update their prototype information. It interacts with the DataOracle contract, which, acting as an oracle, provides off-chain data, extending its capabilities and enabling it to interact with real-world information.

The PrototypeBuffer contract incorporates several methods and functions to enable various operations. The \_addPrototype function is responsible for assigning a new or updated prototype to an owner in the system. The function first checks whether the caller is an authorized owner. If they are, the prototype string provided as input (p) is assigned to the owner's address. Additionally, any existing oracle data associated with the owner is cleared to ensure the prototype's context is up-to-date. The \_seeAll function allows an owner to view the prototypes of other nodes in the system. The function collects all prototypes from other nodes (excluding the caller) and returns them as a set. This ensures that owners have visibility into the distributed learning process by being able to see the models (prototypes) shared by their peers.Table~\ref{tab:notation2} details the notation used for these two algorithms. The contract owner can add a new owner through the setOwner function, granting them system access. The contains function verifies if the caller is an owner, ensuring that only authorized users can conduct specific operations. The contract owner can link the PrototypeBuffer contract with the DataOracle contract via the setOracleInstanceAddress() function. Lastly, the updateData() function asks the DataOracle contract to refresh the data stored in the PrototypeBuffer contract.

The contract includes two modifiers: allowedOwner and onlyOracle. The former checks whether the caller is an authorized owner, while the latter ascertains if the caller is the oracle.

The DataOracle contract features a set of methods and functions. The fetchOracleData() function retrieves the most recent data from the oracle for its use by the PrototypeBuffer contract. The updateOracleData() function updates the available information of the system by setting the most recent data in the DataOracle contract. The sendOracleDataToBuffer function enables the DataOracle contract to supply updated data to the PrototypeBuffer contract, ensuring the system can access to the most recent information.

\begin{figure*}
\centering
\begin{tikzpicture}
\node[rectangle split, rectangle split parts=2, draw, text width=0.45\columnwidth+4cm] (PrototypeBuffer) {
  \textbf{PrototypeBuffer}
  \nodepart{second}
  \begin{tabular}{l}
    -owners: address[] \\
    -oracleData: string \\
    -oracleAddress: address\\
    -oracleInstance: OracleInterface \\
    -Owners: mapping(address $->$ bool) \\
    -ownerToPrototype: mapping(address $->$ string) \\
  \end{tabular}
};

\node[below=of PrototypeBuffer, yshift=-0.5cm, rectangle split, rectangle split parts=2, draw, text width=0.45\columnwidth+1cm] (Methods) 
{
  \textbf{Methods}
  \nodepart{second}
  \begin{tabular}{l}
    +\_addPrototype() \\
    +\_seeAll() \\
    +setOwner() \\
    +setOracleAddress() \\
    +updateData() \\
  \end{tabular}
};

\begin{scope}[xshift=8cm]
\node[rectangle split, rectangle split parts=2, draw, text width=4.1cm] (DataOracle) 
{
  \textbf{DataOracle}
  \nodepart{second}
  \begin{tabular}{l}
  -dataValue: string\\
  -oracleOwner: address \\
  \end{tabular}
};

\node[below=of DataOracle, yshift=-1cm, rectangle split, rectangle split parts=2, draw, text width=4.1cm] (Methods2) 
{
  \textbf{Methods}
  \nodepart{second}
  \begin{tabular}{l}
    +fetchOracleData() \\
    +updateOracleData() \\
    +sendOracleDataToBuffer() \\
  \end{tabular}
};
\end{scope}

\draw[-latex] (PrototypeBuffer) -- (Methods);
\draw[-latex] (DataOracle) -- (Methods2);

\draw[-latex] (Methods2.two west) to[out=105, in=0, looseness=1.1] node[midway, above=2 mm, sloped]{sendOracleDataToBuffer()} (Methods.two east);
\end{tikzpicture}

\caption{UML diagram of the PrototypeBuffer and DataOracle smart contracts, illustrating their attributes, methods, and the relationship between them.}
\label{fig:uml}
\end{figure*}

The DataOracle contract sends back data via the sendOracleDataToBuffer() function. The process involves several steps: Firstly, the PrototypeBuffer contract requests data from the DataOracle contract through the updateData() function. In response, the DataOracle contract emits the fetchOracleDataEvent, signaling data demand. Off-chain, the oracle retrieves the necessary data and then sends it back to the PrototypeBuffer contract through the updateOracleData() function. This is accomplished by calling the sendOracleDataToBuffer() function of the PrototypeBuffer contract and emitting the updateOracleDataEvent.

This implementation underscores the value of using oracles to supply off-chain data to smart contracts, thereby bridging the gap between blockchain systems and external data sources. In this work, we emphasizes the pivotal role of oracles in expanding smart contracts' potential use cases, making them more versatile and adaptable to a broad spectrum of applications.

Figure~\ref{fig:uml} details the attributes, methods, and interrelations between these smart contracts, offering a comprehensive overview of their structure and functionality. It also shows the interactions between the contracts, as well as the roles and responsibilities of each contract within the larger system.

\subsection{Access Control: Administration, Whitelisting, and Permissionless Mechanisms}

The blockchain system we have implemented, evidenced by the supplied Solidity files, utilizes a strategic mixture of administrative control, whitelisting, and permissionless processes to enforce secure and monitored access to contract functions. By judiciously regulating these access control elements, the system can uphold the integrity of stored data, while concurrently fostering widespread participation and engagement with the oracle data. Striking this balance is paramount to the success and security of any blockchain-oriented platform.

A comprehensive technical review of the two smart contracts, namely PrototypeBuffer and DataOracle, illuminates the roles these access control mechanisms play, and their invaluable contributions towards the overall security and effective operation of the system.

Primarily, the system's administration hinges on the Ownable contract sourced from the OpenZeppelin library. Both PrototypeBuffer and DataOracle contracts inherit from this parent contract. The Ownable contract introduces a simple yet robust mechanism for overseeing contract ownership, thus empowering the contract owner with the ability to execute specific privileged operations.

The owner of the contract possesses the authority to include new prototype owners into the whitelist (within PrototypeBuffer) and alter the data source for the oracle (in DataOracle). This administrative role is critical for managing access control, thereby ensuring the smooth operation of the system.

Through the Owners mapping and the allowedOwner modifier, the PrototypeBuffer contract employs a whitelisting process. This setup allows the contract owner to delegate access to selected users (prototype owners) by integrating their addresses into the whitelist.

The whitelisted prototype owners can execute tasks such as adding or modifying their prototypes, as well as accessing all prototypes along with their respective oracle data. This exclusive access control warrants that only authorized individuals can engage with the contract, thereby deterring unauthorized entities from meddling with the stored data.

In contrast, the DataOracle contract operates on a largely permissionless basis. The fetchOracleData() function is publicly accessible, thus enabling any user to request data from the oracle. This architectural decision enhances the accessibility of the oracle data, allowing for wider participation and application of the oracle-provided data.

However, the updateOracleData() function in DataOracle is exclusive to the contract owner, ensuring that only the trusted oracle can supply updated data to the PrototypeBuffer contract. This limitation sustains the integrity and credibility of the oracle data, thwarting unauthorized entities from embedding fraudulent or harmful data into the system.

In conclusion, the access control mechanisms embedded within the PrototypeBuffer and DataOracle contracts strike an optimal balance between security and usability, establishing an effective administration, whitelisting, and permissionless infrastructure. This design cultivates a secure, controlled environment while simultaneously encouraging widespread engagement and interaction with the oracle data, thereby contributing significantly to the success of the blockchain-driven system.

\subsection{SwiftAgg+}

In the proposed architecture, SwiftAgg+ protocol is included to secure the data aggregation process. After the implementation of Shamir's Secret Sharing~\cite{shamir1979share} and MDS erasure coding~\cite{kasami1984probability}, the nodes collectively aggregate the data using the polynomial function: $f(x) = a_0 + a_1 x + a_2 x^2 + \cdots + a_{n-1} x^{n-1}$. This function processes the encoded shares of data from each node to deliver a single aggregate value corresponding to each sensor. 


The final phase uses this securely aggregated data to train the machine learning model and also to store this model in the distributed ledger. Each authorized party is issued a unique private key to sign the transactions on the ledger. To retrieve the data, each party must provide a valid digital signature generated from its private key, which is then verified using the corresponding public key.

In summary, the integration of the SwiftAgg+ protocol in our architecture provides an advanced and secure technique for the aggregation of data. Its implementation of cryptographic techniques, MDS erasure coding, and the distributed ledger formulates a robust framework ensuring data integrity and privacy.

\section{Implementation, Results, and Discussion}
\label{sec:results}

In this section, we present the results of our experimental evaluation of the proposed system, discuss the implementation details, and provide a comprehensive discussion on the findings. With this aim, we organize this section as follows. First, we describe the datasets (Section \ref{sec:datasets}) and the evaluation settings (Section \ref{sec:implementation_settings}) to detail: (i) the hardware use for sensors and nodes; (ii) the software use to integrate the machine learning and the blockchain technologies; and (iii) the metrics to assess the performance. After that, we detail the four tests we have performed to check our proposal: (i) an accuracy test that compares the performance of the decentralized machine learning approach using the XuILVQ algorithm with and without blockchain technology (Section \ref{sec:baselineTest}); (ii) tests that focus on assessing the efficiency of the system (Section \ref{sec:efficiencyTest}) taking into account two parameters: checking the impact of blockchain on the training time on the same decentralized machine learning scheme and assessing the gas consumption within the blockchain layer; finally (iv) a test to check the resilience of our proposal against adversarial attacks (Section \ref{sec:adversarialTest}).

\subsection{Privacy considerations and alternative technologies}

Our approach focused on sharing model updates as prototypes rather than raw data. Since XuILVQ is an incremental learning algorithm, these prototypes are abstracted representations of the data, minimizing privacy risks. While integrating differential privacy could further secure these updates, it would introduce unnecessary computational overhead, given the lightweight nature of the prototypes. For this reason, we chose not to apply differential privacy in our design.

Similarly, IPFS (InterPlanetary File System) was considered for storing and referencing the prototypes. However, after evaluating the size of the prototypes, we determined that a direct blockchain upload was more efficient. The small size of the prototype updates, comparable to the size of the IPFS hash itself, makes IPFS redundant in this context. Moreover, integrating IPFS would introduce additional communication overhead, increasing latency. By directly uploading prototypes to the blockchain, we minimized communication steps and resource consumption.


\subsection{Datasets}
\label{sec:datasets}

With the aim of assessing the performance of our approach, we conducted several tests to evaluate its accuracy, training time, and memory usage. For the experimental evaluation we have used the River\footnote{https://riverml.xyz/}~\cite{montiel2021river} datasets, ImageSegments, Phishing and Bananas.

\paragraph{ImageSegments~\cite{imagesegments}} This dataset consists of $2{\small,}310$ instances of image segment data derived from $3 \times 3$ spatial neighborhoods of outdoor images. Each instance contains $19$ attributes that characterize the image segments, such as average brightness, color, saturation, and hue. This dataset also includes a class label for each instance, categorizing the segments into one of the seven classes: brickface, sky, foliage, cement, window, path, and grass. The ImageSegments dataset is commonly used for classification tasks, pattern recognition, and machine learning algorithm evaluation, particularly focusing on image segmentation and analysis.

\paragraph{Phishing~\cite{phishing}} This is a binary classification dataset designed to identify phishing websites, consisting of $1{\small,}250$ instances. Each instance, which represents a website URL, has 30 attributes that describe characteristics such as the presence of SSL certificates, the URL length, the domain registration length, and the use of pop-up windows, among others. Each instance is labeled as either legitimate or phishing, indicating whether the website is a phishing attempt or not. The Phishing dataset is widely used for evaluating machine learning algorithms, particularly in the context of cybersecurity, to develop models that can effectively detect and prevent phishing attacks.

\paragraph{Bananas~\cite{bananas}} This is a synthetic, two-dimensional dataset designed for binary classification tasks, consisting of $5{\small,}300$ instances. Each instance has two continuous features representing the $x$ and $y$ coordinates of a point in a Cartesian plane. The instances are labeled as either class $1$ or class $2$, resembling the shape of two intertwined bananas. The dataset is particularly useful to evaluate the performance of machine learning algorithms, especially those dealing with non-linear decision boundaries. Additionally, the Bananas dataset serves as an excellent tool to visualize and understand the behavior of classification techniques, such as support vector machines, decision trees, and neural networks.

\subsection{Implementation settings}
\label{sec:implementation_settings}

\paragraph{Algorithm Selection and Configuration}
As it was previously mentioned, we use the Incremental Learning Vector Quantization (XuILVQ) algorithm to perform the machine learning tasks (Section~\ref{sec:background-il}). XuILVQ is robust and is able to properly manage high-dimensional and large-scale datasets, typical in IoT scenarios. The XuILVQ algorithm is particularly appealing because of its ability to adaptively update learning models in response to new data. The initial algorithm parameters for the performed experiments are alpha\_winner=0.9 and alpha\_runner=0.1, which were dynamically adapted across the simulation.

\paragraph{Hardware Setup}

The experiments were conducted on a mixed hardware environment to closely emulate real-world IoT scenarios. 
\begin{itemize}
\item [-] Raspberry Pi Zero W devices were selected as sensors: the elements that collect data. They have lower power consumption than the regular Raspberry Pi line of products, but they have an adequate processing capability to collect, handle and transmit data in the scenario summarized in Figure \ref{fig:network}. They were configured to operate at variable sampling frequencies, depending on the experimental needs, ranging from 1~Hz to 10~Hz.
\item [-] Raspberry Pi 4 Model B devices were selected as computation and blockchain nodes in Figure \ref{fig:network}. They are equipped with a 64-bit quad-core ARM Cortex-A72 CPU (BCM2711) clocked at 1.8 GHz and 4GB LPDDR4-3200 SDRAM of RAM, so they are able to handle more intensive computations, such as the model training. 

\item [-] Finally, a PC featuring an Intel i7-9700K CPU, 32 GB of RAM, and an NVIDIA RTX 2080 Ti GPU was selected as the element to evaluate the machine learning models.
\end{itemize}


\paragraph{Software and Libraries}
The system implementation uses Python 3.8 for its wide support of scientific computing libraries and its advantages to be integrated with IoT devices. We highlight here some of the most relevant libraries:
\begin{itemize}
\item [-] Scikit-learn \cite{kramer2016scikit} for baseline machine learning algorithms and model evaluation tools.
\item [-] Pandas \cite{mckinney2011pandas} and NumPy \cite{bressert2012scipy} for data manipulation and numerical computations, ensuring efficient handling of large datasets.
\item [-] River \cite{montiel2021river} for online machine learning, allowing incremental model updates directly from streaming data.
\item [-] Web3.py to interface with Ethereum blockchain, managing smart contracts and transactions.
\end{itemize}

\paragraph{Blockchain Integration and Smart Contract Development}
Ethereum \cite{dannen2017introducing} smart contracts were deployed to handle data validation, aggregation, and model updates securely. These contracts were developed in Solidity. The Remix IDE \cite{amir2020remix} was also utilized for smaller, iterative testing and debugging of smart contracts. Key features of the contracts included ownership management (using the OpenZeppelin's Ownable contract), transaction limit controls to optimize blockchain resource usage, and secure prototype update mechanisms.

\paragraph{System Optimization and Performance Monitoring}
As it was detailed in the proposal (Section \ref{sec:system}), only those prototypes that shows a statistically significant improvement (defined by a 5\% confidence interval) are uploaded and shared through a blockchain transaction. This decision reduces the number of blockchain interactions and, consequently, the system latency and resources consumption. However, and in order to check this decision is adequate, we need to check and compare its performance with a traditional federated learning (without this limit).


\paragraph{Evaluation Metrics}
The performance of the proposal is evaluated based on several metrics, including model accuracy, training time, memory usage, and blockchain overhead. This selection is grounded on studies (\cite{DesignSurvey}) that identify optimal designs for various tests aimed at analyzing system performance in blockchain integration.

\subsection{Accuracy Test}
\label{sec:baselineTest}
The objective of this first set of tests focuses on the accuracy of the global proposal: the decentralized machine learning approach with the blockchain technology. With this aim, we use the three datasets described in Section \ref{sec:datasets} under three different configurations:


\begin{itemize}
    \item [-] Centralized. The algorithm was implemented in a traditional centralized data processing environment where all computations were conducted on a single server.
    \item [-] Decentralized (Federated) Learning. The model was distributed across four nodes, each one contributes to the learning process in a coordinated federated environment without transaction limits, with continuous updates.
    \item [-] Decentralized (Federated) Learning with Limited Transactions. It is similar to the previous scenario, but with a constraint on the update frequency, allowing only significant model updates to contribute to the global model, with the aim of reducing noise and overfitting.
\end{itemize}

For each one of these three configurations, we calculated the classification accuracy, as the ratio of correctly predicted instances to total instance, that we represent as a percentage.  The data from the datasets are divided between the nodes in a non-iid way, every client has the same number of samples but the class distributions are imbalanced, with clients having varying proportions of samples from each class. The results are summarized in Figures~\ref{fig:AccuracyTest1}, \ref{fig:AccuracyTest2} and \ref{fig:AccuracyTest3}:


\begin{figure}[t]
  \centering
  \includegraphics[width=0.5\linewidth]{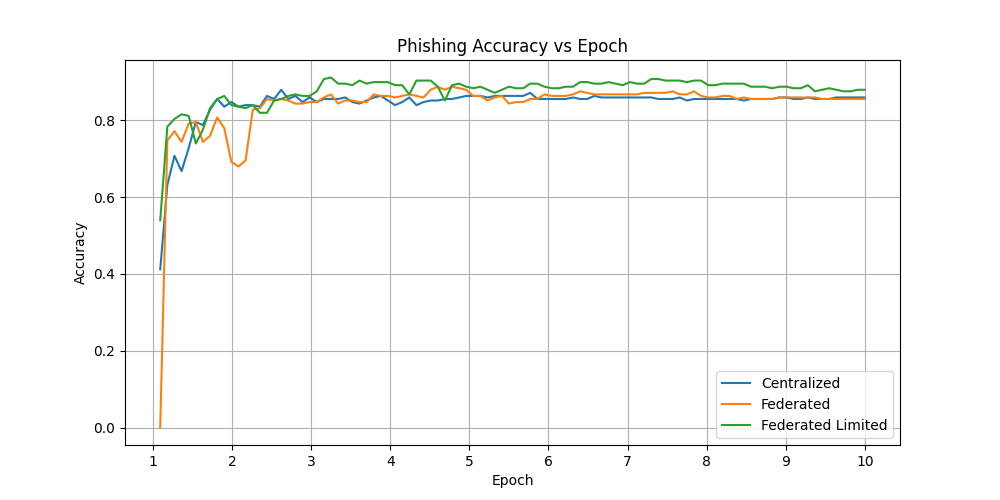}
  \caption{Phishing dataset performance.}
  \label{fig:AccuracyTest1}
  \includegraphics[width=0.5\linewidth]{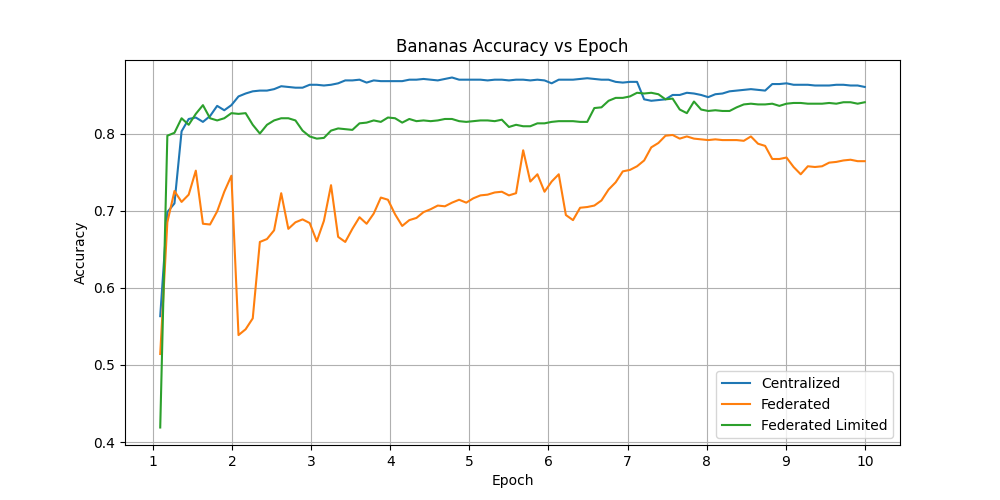}
  \caption{Bananas dataset performance.}
  \label{fig:AccuracyTest2}
  \includegraphics[width=0.5\linewidth]{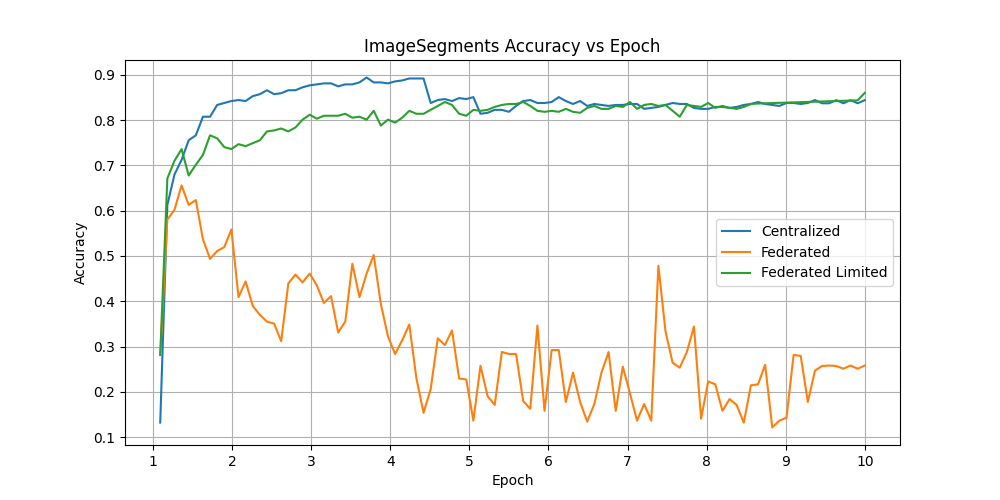}
  \caption{ImageSegments dataset performance.}
  \label{fig:AccuracyTest3}
\end{figure}

We can observe that the centralized learning configuration, as expected, shows the highest accuracy for all datasets. Thus, it provides a steady and consistent learning progression with few fluctuations, which offers a suitable benchmark for the decentralized algorithms. The second scenario, decentralized (federated) learning shows more variability in terms of accuracy, particularly with the ImageSegments and, in a minor way, the Bananas dataset. This indicates potential challenges in model synchronization and in the uniform distribution of data across nodes. Finally, the decentralized (federated) learning setting with limited transactions generally provides better performance results, compared to the unlimited transactions scenario.

\subsubsection{Data Heterogeneity Across Nodes}
One significant challenge in federated learning is how data is distributed across nodes. In our setup, each node received the same amount of data, but the data was unevenly distributed in terms of class balance. This uneven distribution can cause some nodes to have more data from certain classes than others, leading to imbalanced training.
\begin{itemize}
    \item The Bananas dataset consists of two intertwined clusters that resemble the shape of bananas. When these clusters are split across nodes, each node may only see part of the overall pattern. This incomplete view makes it harder for the global model to learn the full data structure, resulting in lower accuracy in the Federated configuration.
    
    \item Similarly, in the ImageSegments dataset, the different classes (e.g., brickface, sky, foliage) were not evenly distributed across nodes. Some nodes were more focused on certain classes than others. This imbalance caused a drop in the performance of the federated model because it didn’t have the benefit of seeing the entire dataset at once, as would be the case in a centralized learning model.
\end{itemize}

\subsubsection{Dataset Complexity}
Both the Bananas and ImageSegments datasets introduce additional complexity that further challenges the federated learning approach.
\begin{itemize}
    \item The distinctive cluster shapes in the Bananas dataset make it difficult for the model to fully capture the global data structure.
    
    \item For ImageSegments, the high-dimensional features, such as color intensity and texture metrics, are challenging for the federated model to capture when distributed across nodes. In the centralized setup, where the model has access to the entire dataset, it can better learn and generalize these complex features, leading to higher accuracy.
\end{itemize}

\subsubsection{Selective Prototype Sharing in Limited Transactions}
In the \textit{"Federated with Limited Transactions"} configuration, we saw better performance compared to the standard Federated setup. This is because only the most useful model updates were shared between nodes, while less useful updates were filtered out. This approach helped reduce the noise in the model updates, resulting in a cleaner and more efficient learning process:
\begin{itemize}
    \item Reduction of Noisy and Poor-Quality Update: By limiting the number of updates shared between nodes, we avoided sending low-quality updates that might have negatively impacted the global model’s accuracy.
    
    \item Avoiding Overfitting by Reducing Redundant Information: Fewer but more meaningful updates helped the model avoid overfitting to local data patterns, leading to better generalization across the whole dataset.
\end{itemize}

The Phishing dataset, on the other hand, didn’t show the same level of improvement in the "Federated with Limited Transactions" configuration. This is because the Phishing dataset is simpler: it’s a binary classification task (phishing vs. legitimate), and the features are more uniform. Therefore, the benefits of limiting updates are less noticeable in this case.

\begin{figure}[h]
  \centering
  \includegraphics[width=\linewidth]{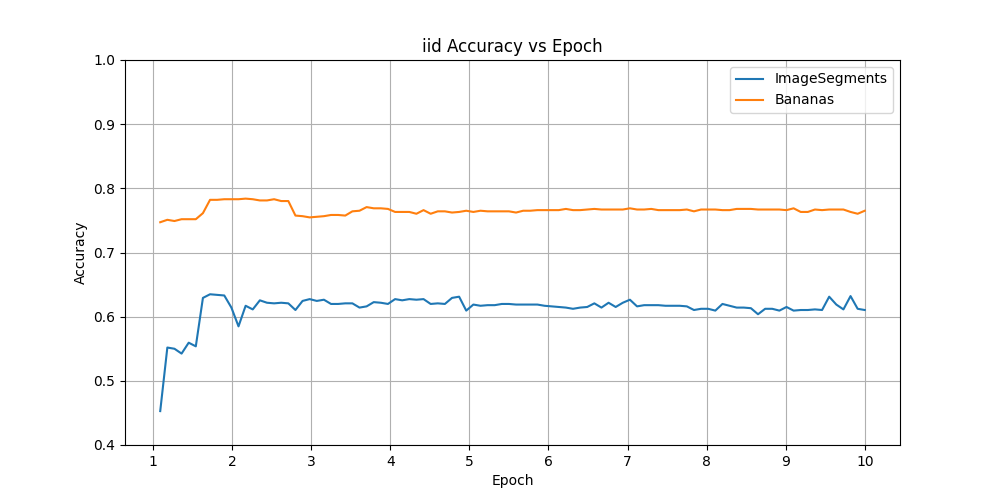}
  \caption{Accuracy over epochs for iid in Bananas and ImageSegments dataset.}
  \label{fig:accuracy_drop_comparison}
\end{figure}

To further understand how data distribution affects performance, we ran a test where the data was more evenly distributed across nodes (i.e., all nodes had a more balanced mix of classes) in Figure~\ref{fig:accuracy_drop_comparison}.

For the Bananas dataset, we observed a small improvement in the stability of the model during training, but the final accuracy remained similar to the previous results. In contrast, the ImageSegments dataset shows a more significant improvement in both the final accuracy and the convergence rate. This indicates that the heterogeneity of data, which negatively impacted performance in traditional federated learning, does not pose the same challenge in our "Federated with Limited Transactions" configuration. By selectively sharing the top 5\% of high-quality prototypes, our approach mitigates the issues caused by non-i.i.d. data across nodes, leading to better model convergence and generalization.

\subsection{Efficiency tests}
\label{sec:efficiencyTest}

After establishing a fundamental understanding of how blockchain integration and the implementation of selection-based prototypes improve performance in federated learning environments, we assessed the system efficiency and effectiveness. Being more specific, we focuses on obtaining the training time (Section \ref{sec:trainingTimeTest}) and also the gas costs related to transactions (Section \ref{sec:gascost}). Both parameters are crucial because they directly impact the scalability and economic sustainability of machine learning models in decentralized schemes. By quantifying the computational overhead and cost repercussions, we would provide a complete assessment of the trade-offs associated with using blockchain technology in federated learning settings.

We are going to test both parameters under two scenarios: unlimited transactions and limited transactions. The former, the traditional one, allows nodes to share the information of the local models (all prototypes in our algorithm) with all the nodes in the learning network at each round. The latter, which is one of the key aspects of our proposal, limits this sharing scheme as it was previously described (Section \ref{sec:algorithmic_approach}): only prototypes that show significant improves in the performance ($5\%$ performance interval) are shared.

\subsubsection{Training Time and Communication Overhead Test}
\label{sec:trainingTimeTest}


\begin{table*}[t]
\centering
\begin{tabular}{|l|c|c|c|}
\hline
\textbf{Learning Model} & \textbf{Time/Epoch (s)} & \textbf{Transactions/Node} & \textbf{Overhead (MB)/ Epoch} \\ \hline
FL (Standard) & 46.74 & N/A & N/A \\ \hline
Blockchain FL (Limited) & 47.91 & 15.5 & 0.12 \\ \hline
Blockchain FL (Unlimited) & 88.3 & 250 & 2.06 \\ \hline
\end{tabular}
\caption{Training times, transactions per node, and communication overhead per model in KB.}
\label{tab:training_times}
\end{table*}

The objective of this section is to evaluate the impact of blockchain on both training time and communication overhead in federated learning systems, assessing if the security benefits justify the additional computation time and communication costs. To do this, we measure the training time from initialization to convergence (including total training time and time per epoch) as well as the communication overhead in terms of the data exchanged between nodes. All the results were obtained using the Phishing dataset.

 Table~\ref{tab:training_times} summarizes the results, showing both training times and the communication overhead per model in terms of the total number of transactions and the corresponding data size transmitted per node.

The former, which is the traditional blockchain integration approach, increases the epoch times in line with previous research results~\cite{TrainingTime}: nearly duplicate the effective time of the total training time. However, with the approach where transactions are limited, we observe a negligible increase of only $1$ second in the total training time, also reducing the number of transactions used.

 To put this in context, each transaction stores 2104 bytes (approximately 0.002 MB), meaning that the communication overhead for the limited transactions configuration is only 0.03117 MB per node compared to 0.50163 MB for the unlimited transactions case. This reduction in communication overhead, along with the minor increase in training time, makes the limited transactions approach much more efficient, especially in environments where network bandwidth and latency are critical concerns.

To sum up, blockchain integration increases the training time, but this overhead is manageable within real-world environments, especially in applications where high levels of security and data integrity are crucial. The benefits provided by blockchain, such as improved security attributes, justify the additional computational time.

\subsubsection{Blockchain Gas Cost Test}
\label{sec:gascost}

The second objective is to evaluate the impact of transaction limits on gas consumption and system accuracy within our scenario: a federated learning framework integrated with blockchain technology.

Table \ref{tab:costs} shows the cost in ETH\footnote{Ether (ETH) is the native cryptocurrency of the Ethereum platform. It serves multiple purposes within this ecosystem, such as paying for gas fees (interacting with the network) and executing smart contracts.} \footnote{Gwei represents one billionth of an Ether ($10^{-9}$ ETH). The term "gwei" is derived from "giga-wei," where "wei" is the smallest unit of Ether. Gwei is commonly used to specify transaction fees (gas prices) on the Ethereum network due to its convenient size for such purposes.} for each category in the contract. According to this information, we can obtain the total deployment cost of our contracts, which is 0.02596727 ETH. This figure showcases the efficiency of our solutions, since it is lower than other contract deployment costs in the literature, such as the 0.21600988 ETH~\cite{gasPrice} or 0.04746 ETH~\cite{gasPrice2}.


\begin{table}[H]
\centering
\begin{tabular}{|c|c|c|}
\hline
\textbf{Component} & \textbf{Gas Used} & \textbf{Cost in ETH} \\ \hline
Security & 88,546 & 0.001063 \\ \hline
Oracle & 139,731 & 0.001677 \\ \hline
I/O & 656,230 & 0.007875 \\ \hline
\end{tabular}
\caption{Ethereum costs for contract components at 12 Gwei.}
\label{tab:costs}
\end{table}


\subsection{Comparison with other Implementation}

In order to compare we implemented a modified version of FedProto \cite{fedProto}, a well-regarded federated learning algorithm. FedProto typically shares averaged class prototypes across clients, reducing the amount of data exchanged to protect privacy. We adapted this framework to support incremental learning by using XuILVQ as the internal model, which allows clients to update and communicate only relevant prototypes during each round.

The purpose of this experiment was to compare our architecture with an established federated learning architecture to measure its performance in terms of accuracy and communication efficiency, particularly when adapted for incremental learning scenarios. This comparison helps assess the strengths of our system, especially regarding the selective prototype-sharing feature.

Using the modified FedProto configuration, we achieved a mean accuracy of 79.1\% on the Phishing dataset. This result is lower than the 84.4\% accuracy attained with our XuILVQ-based federated learning system, even without the added benefit of our prototype-sharing optimization. Moreover, the modified FedProto implementation incurred a higher communication overhead of 0.15 MB, compared to 0.03117 MB for our proposed method.

This comparison underscores that our approach provides a more efficient solution for federated incremental learning, lowering communication overhead while enhancing accuracy. The advantages of our architecture are particularly pronounced in IoT applications that require efficient data transmission and real-time model updates.

\subsection{Adversarial Robustness Test}
\label{sec:adversarialTest}

Finally, we tested the resilience of our federated learning model against two types of adversarial attacks: data poisoning and model inversion. For the data poisoning scenario, we configured a federated learning environment setting of 10 nodes. Then, we randomly selected some of them to be poisoned nodes, that is, nodes that alter the data received through total label flipping to simulate data poisoning attacks. The impact of these attacks on the overall system performance was measured assessing the degradation of the model accuracy.


Figure~\ref{tab:accuracy_drop_comparison} compares the impact on the model accuracy when we have $10\%$, $20\%$, $30\%$, $50\%$, $70\%$, $90\%$ and $100\%$ of malicious nodes to the results when there are no poisoned nodes.

\begin{table}[t]
\centering
\begin{tabular}{|c|c|}
\hline
\textbf{\% malicious nodes} & \textbf{Final Accuracy}\\ \hline
        0 & 84.4\\ \hline
        10 & 79.6\\ \hline
        20 & 71.2\\ \hline
        30 & 62.8\\ \hline
        50 & 50.2\\ \hline
        70 & 43.2\\ \hline
        90 & 20.2\\ \hline
        100 & 14.8\\ \hline
\end{tabular}
\caption{Accuracy over epochs from different percentage of malicious nodes}
\label{tab:accuracy_drop_comparison}
\end{table}

Additionally, we implemented a model inversion attack to evaluate the privacy protection performance of our system \cite{heterogeneity}. This attack attempts to reconstruct the original feature data from the model updates. We applied this attack to the Phishing dataset, where features are normalized between 0 and 1. The success of the inversion was measured using the Mean Squared Error (MSE) between the reconstructed and original data, with a lower MSE indicating a more successful attack.

With an MSE threshold of 0.1, the attack was only able to generate one close reconstruction (MSE = 0.09) in the fl architecture, which still did not reach the stricter threshold of 0.01, typically required for a successful inversion in privacy-sensitive applications; and 0 in the limited transactions architecture. The overall success rate of the attack was extremely low, demonstrating the model's robustness to inversion attempts.

These results indicate that the proposed method is not only resistant to data poisoning attacks but also robust against model inversion attacks, ensuring strong privacy protection in federated learning environments.

\section{Conclusions and Future Work}
\label{sec:con}

Our proposal defines a Blockchained-Federated Learning solution that integrates the incremental learning vector quantization algorithm (XuILVQ) with Ethereum blockchain technology, addressing the challenges of data privacy, security, and scalability in decentralized machine learning environments. This scheme provides a secure and efficient framework for data sharing, model training, and prototype storage in a distributed setting.

The combination of the previously mentioned technologies are specially suitable for IoT environments, where different computation nodes with limited computation resources must adaptability learn from streams of data in a decentralized scheme. After performing different tests, we can conclude first that our proposal ensures the integrity and confidentiality of data during transmission between nodes. Besides, it also improves accuracy of the model compared to other previous approaches in the literature. Finally, our scheme facilitates an efficient model training and prototype storage, which ultimately reduces communication overheads without penalizing the model accuracy. Consequently, the integration of XuILVQ decentralized learning algorithm and Ethereum blockchain technology in our solution offers a promising approach to address the ongoing challenges of collaborative machine learning in IoT settings.


However, there are some interesting aspects to explore, such as add new optimized consensus mechanisms able to reduce, even more, the inherent overheads for communication in decentralized learning systems. In fact, we are currently working on this area using a reinforce learning mechanism that adds new conditions to share the intermediate models among the computation nodes. Additionally, we are also facing how to detect malicious nodes and how to design quarantine procedures for them that reduce the influence of wrong data to build the global learning model.

For future work, we plan to explore the integration of IPFS in scenarios where model updates are larger, and communication overhead becomes a more significant factor. In ongoing projects, IPFS has proven to be more effective for managing large data messages. This technology may be relevant for systems handling more complex data models, where efficiency in data exchange is crucial.





\section*{Acknowledgments}

This work was supported by the grant PID2020-113795RB-C33 funded by MICIU/AEI/10.13039/501100011033 (COMPROMISE project); by the Spanish Government under these research projects funded by MCIN/AEI/10.13039/501100011033: DISCOVERY PID2023-148716OB-C32; the grant "TRUFFLES: TRUsted Framework for Federated LEarning Systems", within the strategic cybersecurity projects (INCIBE, Spain), funded by the Recovery, Transformation and Resilience Plan (European Union, Next Generation)" and by the Axencia Galega de Innovación (GAIN) (25/IN606D/2021/2612348). Additionally, this work has also received financial support from the Galician Regional Government under project ED431B 2024/41 (GPC).

\bibliographystyle{ieeetr}
\bibliography{biblio}

\end{document}